# Low-Back Pain Physical Rehabilitation by Movement Analysis in Clinical Trial

Sao Mai Nguyen [1]

[1] U2IS, ENSTA / IP Paris, France.

Email: nguyensmai@gmail.com

## INTRODUCTION

To allow the development and assessment of physical rehabilitation by an intelligent tutoring system, we propose a medical dataset of clinical patients carrying out low back-pain rehabilitation exercises and benchmark on state of the art human movement analysis algorithms.

This dataset is valuable because it includes rehabilitation motions in a clinical setting with patients in their rehabilitation program. This paper introduces the **Keraal dataset**, a clinically collected dataset to enable intelligent tutoring systems (ITS) for rehabilitation. It addresses **four challenges** in exercise monitoring:

1. **Motion assessment** – is the exercise performed correctly?
2. **Error recognition** – what type of error occurred?
3. **Spatial localization** – which body part is responsible?
4. **Temporal localization** – when did the error occur?

## RELATED WORK

Several datasets have advanced human activity recognition, including those built with Kinect, wearable sensors, or multi-camera setups. Yet few were designed for rehabilitation. Early datasets such as K3Da[3] and HPTE[6] collected data in healthcare contexts but with healthy volunteers and limited annotations. EmoPain[5] focused on pain expression rather than exercise performance, while the Kimore[4] dataset is closest to our work, targeting low-back pain with medical labels. However, Kimore does not provide temporal error localization.

In contrast, the Keraal dataset [9] is the first to capture rehabilitation patients in real clinical programs with comprehensive annotations for all four challenges.

## DATASET AND METHODOLOGY

### Participants and Exercises

The dataset was collected during a four-week rehabilitation program involving 31 patients in a double blind clinical study to investigate the feasibility of a robot coach analysing patient's rehabilitation movement using a camera. 12 patients were randomly selected to have a daily session with a humanoid robot coach that can show them the physical rehabilitation movements both visually and orally, and can give a feedback whether their movement is correct, using a Kinect. The clinical study showed non-inferiority of a robot coaching program compared to the standard care program, and good user (both patients and therapists) assessment for similar robot coaches for home use. [7]

This dataset includes recordings from 12 patients with chronic low-back pain and 9 healthy volunteers for reference data. Three exercises commonly prescribed for spine rehabilitation were selected: **torso rotations**, **flank stretches**, and a **breathing exercise** with flexed arms. These exercises were chosen because they are representative of physiotherapy practice and can be visually assessed by an intelligent system.

### Sensors and Recordings

Data collection used the Microsoft Kinect V2 to capture RGB videos, depth maps, and 3D skeletal joint positions. We also processed the RGB videos to extract 2D keypoints with OpenPose and BlazePose, and for some sessions, high-precision Vicon motion capture was used as a reference. In [11], we showed that, on average, results obtained through Kinect, OpenPose and BlazePose data were quite comparable.

In total, the dataset includes **2622 recordings**, with 1881 from patients and 741 from healthy subjects.

### Annotation Protocol

Annotations were carried out by a physiotherapist and a physician using the Anvil tool. Movements were labeled at three levels:

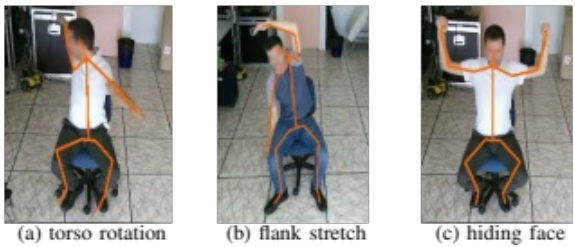

Figure 1 The three rehabilitation exercises in [9]

Table 1: Content of the Keraal Dataset. Group 1A= patients annotated, group 1B =patients without annotation, group 2A - healthy subjects annotated, group 2B= healthy subjects without annotation; group 3 =healthy subjects with Vicon recording

| Gr. | Annotation | RGB videos | Kinect | Openpose/Blazepose | Vicon | Nb rec |
|---|---|---|---|---|---|---|
| 1a | xml anvil : err label, bodypart, timespan | mp4, 480x360 | tabular | dictionary | NA | 249 |
| 1b | NA | mp4, 480x360 | tabular | dictionary | NA | 1631 |
| 2a | xml anvil : err label, bodypart, timespan | mp4, 480x360 | tabular | dictionary | NA | 51 |
| 2b | NA | mp4, 480x360 | tabular | dictionary | NA | 151 |
| 3 | error label | avi, 960x544 | tabular | dictionary | tabular | 540 |

- **Global assessment** (correct, incorrect, incomplete, or motionless),
- **Error classification** (error type, severity, and responsible body part),
- **Temporal localization** (exact start and end frames of the error).

Inter-annotator agreement analysis showed substantial reliability, confirming annotation consistency: Cohen's κ = 0.63 and Krippendorff's α=0.62 [1], [2].

Table 1 summarises in the Keraal dataset the available recording modalities and the format of the data per group.

## BENCHMARKS
To evaluate the dataset, two baseline algorithms were tested: a **Gaussian Mixture Model (GMM)** on a Riemannian manifold [8], a **Long Short-Term Memory (LSTM)** network, and **Hyperformer**, composed of a hyper-graphs and ten laoyers of self-attention [10]. See other algorithms in [13,14].

For **global assessment**, Fig. 2 shows the classification results between correct and incorrect movements for each exercise for the best F1-score of the GMM baseline and LSTM baseline, respectively. While colors represent normalized values, we left absolute values within the matrices so as to emphasize that classes are imbalanced.. Both GMM and LSTM struggled to detect incorrect movements, with many errors misclassified as correct. The LSTM model performed better than GMM but still showed significant limitations.

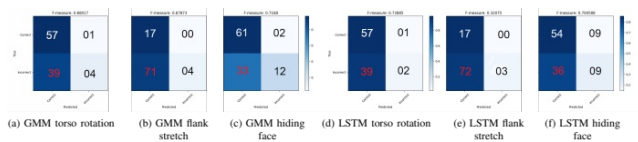

Figure 2: Confusion matrix of error detection using SVM and LSTM autoencoder

For **error classification**, our goal is to classify a performed movement into one of the identified errors. Table 2 shows that the LSTM achieved higher accuracy than GMM with SVM, reaching up to 64% accuracy for torso rotation. However, distinguishing subtle errors, such as insufficient torso rotation or incomplete arm movement, remained challenging Hyperformer perfoms the best. These results demonstrate that standard methods are insufficient and highlight the need for more specialized rehabilitation-focused models.

| Exercise | Hyperformer [11] | LSTM Best | LSTM mean | GMM [15] |
|---|---|---|---|---|
| Torso rotation | 73,17 | 64.44 | 53.89 | 27.78 |
| Flank stretch | 64.10 | 43.04 | 31.64 | 25.32 |
| Hiding face | 74.28 | 56.19 | 49.1 | 33.33 |

**Table 2:** Motion error classification accuracies of GMM-based features combined with SVM classifier. LSTM (best model and mean of several models) and Hyperformer models.

## CONCLUSION AND PERSPECTIVES
The Keraal dataset provides the first clinically grounded benchmark for rehabilitation exercise analysis. It includes multimodal recordings, patient data, and detailed annotations covering correctness, error types, and spatio-temporal details. While the dataset is limited to three exercises and a relatively small number of patients, it represents a significant step toward enabling intelligent tutoring systems that can autonomously monitor rehabilitation.

Future work will extend the dataset to include more exercises and larger participant groups. Algorithmically, research should focus on models capable of fine-grained error detection and robust handling of patient variability. The dataset also supports broader applications in computer vision, robotics, biomechanics, and telemedicine.

By enabling automated supervision of exercises, the Keraal dataset contributes to improving patient adherence and access to rehabilitation, especially for those unable to attend clinical sessions regularly. In the long run, it offers the foundation for more effective, accessible, and personalized rehabilitation care [12], especially with intelligent tutoring systems stimulating their intrinsic motivation through interactive learning [16].

## ACKNOWLEDGEMENTS
This work is partially supported by the EU FP7 grant ECHORD++ KERAAL and by the European Regional Fund via the VITAAL Contrat Plan Etat Region, and of the Hi! PARIS Engineering Team.